\documentclass[11pt]{article}
\usepackage[margin=1in]{geometry}
\usepackage[T1]{fontenc}
\usepackage{lmodern}
\usepackage{cite}
\usepackage{amsmath,amssymb,amsfonts}
\usepackage{algorithmic}
\usepackage{algorithm}
\usepackage{graphicx}
\usepackage{multirow}
\usepackage{booktabs}
\usepackage{array}
\usepackage{textcomp}
\usepackage[hidelinks]{hyperref}
\hypersetup{pdftitle={M2PFN: End-to-End Disentangled Alignment for Generalizable Multimodal In-Context Learning in Alzheimer's Disease}}

\begin{document}

\title{M$^2$PFN: End-to-End Disentangled Alignment for Generalizable Multimodal In-Context Learning in Alzheimer's Disease}

\author{%
Lujia Zhong$^{1,2}$, Shuo Huang$^{1,3}$, Jianwei Zhang$^{1,2}$, Xinyu Nie$^{1}$, and Yonggang Shi$^{1,2,3}$\thanks{Corresponding author: Yonggang Shi (\texttt{yshi@loni.usc.edu}).}\\[6pt]
{\small $^{1}$Stevens Neuroimaging and Informatics Institute, Keck School of Medicine,}\\
{\small University of Southern California, Los Angeles, CA 90033, USA}\\
{\small $^{2}$Ming Hsieh Department of Electrical and Computer Engineering, Viterbi School of Engineering,}\\
{\small University of Southern California, Los Angeles, CA 90089, USA}\\
{\small $^{3}$Alfred E. Mann Department of Biomedical Engineering, Viterbi School of Engineering,}\\
{\small University of Southern California, Los Angeles, CA 90089, USA}}
\date{}

\maketitle

\begin{abstract}
While various multimodal methods combining imaging and tabular data for Alzheimer's disease (AD) diagnosis were proposed, they are often limited in generalization across cohorts. In-context learning (ICL) has demonstrated excellent generalization performances and high flexibility in foundational tabular models such as TabPFN. To extend TabPFN's ICL to multimodal AD analysis, the main obstacle is that TabPFN is meta-trained on synthetic tabular priors that do not naturally match the statistical structure of image-derived features. We propose M$^2$PFN, an end-to-end framework that turns this tabular foundation model into a multimodal AD predictor. M$^2$PFN (i) performs differentiable inference through TabPFN's transformer, back-propagating task gradients into 3D-MRI and tabular encoders; (ii) aligns the two modalities into a shared subspace, via disentanglement and a contrastive objective, matched to the ICL engine's prior; and (iii) folds in a frozen tabular-only prediction through a learnable gated shortcut. Because the ICL engine stays frozen, its in-context mechanism is preserved for test-time generalization, while end-to-end training shapes the encoders into features it can exploit. On ADNI ($n=2240$, three-class CN/MCI/AD), M$^2$PFN attains $65.55\%$ macro-F1 and $82.21\%$ macro-AUC, surpassing a comprehensive set of unimodal and multimodal baselines. By swapping only the head for a TabPFN regressor, the same architecture regresses baseline MMSE on a $1250$-subject sub-cohort to test MAE $1.743$, outperforming every multimodal baseline. On two external cohorts (OASIS-3 and SCAN) with no retraining, M$^2$PFN achieves the best AUC and the lowest MMSE MAE across all baselines, and transfers even when the cognitive instrument changes.
\end{abstract}


\section{Introduction}\label{sec:introduction}
For Alzheimer’s disease (AD) diagnosis and prognosis, the value of multimodal analysis that combines complimentary information from imaging and non-imaging tabular data has been well demonstrated~\cite{polsterl2021combining,qiu2022multimodal,yun2024flex,xin2025i2moe,huang2025multistage,yang2025adfound,yang2026taming}. Among imaging modalities, structural MRI is the most widely available and routinely acquired in clinical practice. It characterizes neurodegeneration through cortical and hippocampal atrophy, which are among the most reliable \emph{in vivo} markers of cognitive decline. Non-imaging tabular data, including demographic characteristics, clinical history, genetic factors such as APOE genotype, and fluid biomarkers such as CSF amyloid and tau, provide additional information about disease risk and progression that imaging alone cannot reveal.  Despite their benefits, most existing multimodal approaches combine imaging and tabular data through task- and dataset-specific fusion architectures with fixed input structures. Consequently, they typically require retraining or extensive fine-tuning when transferred across cohorts with different imaging protocols, scanner vendors, and population characteristics, and struggle to accommodate variation in the number and type of available tabular features, ranging from more than one hundred variables in Alzheimer's Disease Neuroimaging Initiative (ADNI)~\cite{mueller2005alzheimer,jack2008alzheimer} to fewer than ten in some external cohorts.

These challenges naturally motivate foundation models, which leverage large-scale pretraining to learn transferable representations that can be adapted across downstream tasks and domains. Such capability is particularly desirable for AD analysis, where both imaging distributions and multimodal feature availability vary considerably across clinical cohorts. However, most existing AD prediction models, whether based on convolutional neural networks~\cite{he2016deep,huang2017densely,hu2018squeeze}, transformers~\cite{dosovitskiy2020image,he2023swinunetr,zhang2022mmformer,tsai2019multimodal}, or mixture-of-experts architectures~\cite{fedus2022switch,jin2024moe++,swamy2024intrinsic,yun2024flex,xin2025i2moe}, remain task-specific, with model parameters optimized for the distributions observed during training. Consequently, they provide limited support for flexible multimodal integration and robust cross-cohort deployment.

\textit{In-context learning} (ICL) offers a principled answer to this foundation-model demand. Rather than compressing data into fixed parameters, a pretrained ICL model conditions on a labeled support set supplied at inference time and predicts a query directly through attention over that support, without updating a single weight. Pioneered by large language models~\cite{brown2020language}, ICL has repeatedly shown strong generalization and rapid adaptation to new tasks and distributions purely by changing the support—a capability structurally unavailable to parametric predictors. For tabular data, this paradigm has already yielded a family of foundation models: prior-data fitted networks (PFNs) such as TabPFN~\cite{hollmann2025accurate} and TabFM~\cite{kong2026tabfm}, meta-trained on large families of synthetic priors, deliver strong, well-calibrated predictions on medium to small-sized datasets typical of clinical cohorts, all off the shelf and with the network frozen. Two obstacles, however, keep this tabular foundation model from multimodal AD analysis. First, no ICL foundation model natively ingests medical imaging or, more generally, multiple modalities; a tabular-only ICL engine therefore leaves most of the AD signal unused. Second, TabPFN is meta-trained on large families of synthetic tabular priors sampled from structural causal models; its learned in-context prior therefore reflects the statistical structure of tabular data-generating processes, not that of image-derived features. Naively extracting image embeddings and feeding them as additional tabular columns yields a representation mismatched to what the engine was trained to exploit, making end-to-end training essential to shape the encoder's output to the engine's prior. However, TabPFN's preprocessing pipeline contains non-differentiable operations that fracture gradient flow, so it cannot be coupled to and jointly trained with the learnable encoder this adaptation demands.

We propose M$^2$PFN, an end-to-end framework that repurposes a frozen, off-the-shelf tabular foundation model for multimodal Alzheimer's disease analysis. M$^2$PFN preserves the frozen TabPFN in-context learning (ICL) engine as the prediction head and learns multimodal representations that are directly optimized for in-context reasoning through end-to-end training. By keeping the ICL engine frozen, M$^2$PFN supports support-set replacement at inference, enabling gradient-free transfer to unseen clinical cohorts without retraining.

\textbf{Contributions.} Our main contributions are summarized as follows:
\begin{itemize}
\item We instantiate differentiable prompt tuning through TabPFN by bypassing its non-differentiable preprocessing pipeline and back-propagating gradients through the in-context attention computation.\footnote{The differentiable-input path for TabPFN's regressor used in this work has been contributed by the authors to the official TabPFN library and merged upstream (\url{https://github.com/PriorLabs/TabPFN/pull/923}).}
\item We introduce a disentangled alignment-based multimodal representation learning strategy that jointly optimizes MRI and tabular features through the frozen ICL engine.
\item We achieve state-of-the-art performance on ADNI for both three-class CN/MCI/AD classification and MMSE regression.
\item We demonstrate robust zero-retraining transfer to two independent external cohorts (OASIS-3 and SCAN) via support-set replacement, achieving the best performance among all gradient-free adaptation baselines.
\end{itemize}

\section{Related Work}\label{sec:related}

\subsection{Multimodal Alzheimer's Disease Analysis}
AD analysis spans two paradigms: categorical classification (e.g., CN/MCI/AD) and regression of continuous cognitive scores such as MoCA and MMSE. Deep learning for both has progressed from unimodal architectures, such as ResNet~\cite{he2016deep}, DenseNet~\cite{huang2017densely}, and transformer-based SwinUNETR~\cite{he2023swinunetr}, to multimodal fusion architectures that combine imaging with clinical/tabular features, spanning cross-modal transformers (MulT~\cite{tsai2019multimodal}, mmFormer~\cite{zhang2022mmformer}) and mixture-of-experts (MoE) routing across modalities (Flex-MoE~\cite{yun2024flex}, MoE++~\cite{jin2024moe++}, SwitchGate~\cite{fedus2022switch}, I2MoE~\cite{xin2025i2moe}, InterpretCC~\cite{swamy2024intrinsic}); these consistently show that combining MRI with tabular/clinical data outperforms either modality alone. All of these methods stay within the parametric learning paradigm, compressing the training distribution into fixed weights and cannot reference labeled exemplars at inference or effectively adapt to a new cohort without retraining or finetuning.

\begin{figure*}[t]
\centering
\includegraphics[width=1\textwidth]{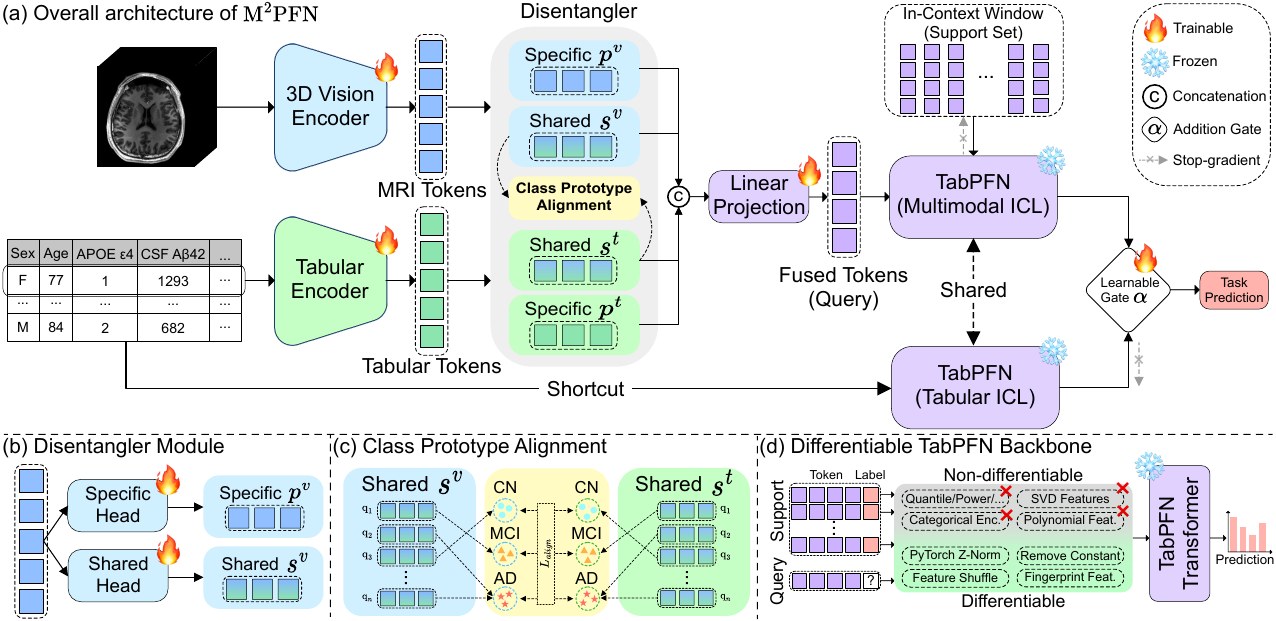}
\caption{Overview of the proposed M$^2$PFN framework (\S\ref{sec:method}). (a) Overall architecture. (b) Disentangler module for separating modality-specific and shared representations. (c) Class prototype alignment for encouraging cross-modal consistency. (d) Differentiable TabPFN backbone enabling end-to-end gradient propagation through the frozen PFN engine.}
\label{fig:overview}
\end{figure*}

\subsection{In-Context Learning}
In-context learning (ICL) was popularized by large language models, which solve new tasks by conditioning on demonstrations without parameter updates~\cite{brown2020language}, and has since been extended to vision-language models~\cite{alayrac2022flamingo}. For tabular data, two main paradigms have emerged: retrieval-based methods that attend to training examples during inference~\cite{kossen2021self,gorishniy2024tabr}, and Prior-Data Fitted Networks (PFNs)~\cite{muller2022transformers}, which meta-learn Bayesian inference over synthetic task distributions. TabPFN~\cite{hollmann2023tabpfn,hollmann2025accurate} is the most mature PFN-based model, achieving competitive performance on small-to-medium tabular datasets without test-time optimization, while TabFM~\cite{kong2026tabfm} further scales this paradigm to larger tabular benchmarks. We build upon TabPFN. However, like existing tabular foundation models, it is designed exclusively for structured tabular inputs and relies on a non-differentiable preprocessing pipeline before the transformer. Consequently, multimodal inputs such as 3D MRI cannot be incorporated into the in-context reasoning process, nor can gradients propagate from the ICL engine back to an imaging encoder for end-to-end representation learning.

\subsection{Prompt Tuning for ICL Engines}
When a large pretrained model cannot be finetuned directly, being expensive to update or easy to collapse on a small dataset, prompt tuning instead optimizes a small set of continuous input vectors to steer the frozen model, spanning soft prompts at the input layer~\cite{lester2021power}, deep/prefix variants that inject tunable vectors at every layer~\cite{li2021prefix,liu2024gpt}, and parameter-efficient adapters such as LoRA~\cite{hu2021lora} that learn a low-rank update to the frozen weights. Prompt tuning a PFN, however, needs a gradient path from the engine's output back to its input, which its non-differentiable preprocessing stack does not provide. A concurrent multimodal PFN, MMPFN~\cite{kim2026multimodalpfn}, works around this architecturally, inserting a multimodal mixer (a multi-head gated MLP with cross-attention pooling) between TabPFN's feature encoder and its transformer and lightly finetuning that mixer on frozen, precomputed image embeddings; this avoids re-extracting image features during training but prevents end-to-end adaptation of the imaging encoder. We take the opposite stance on what to adapt: the engine's weights stay frozen, which prevents a foundation model from collapsing on the small data of a clinical cohort, and task gradients instead flow through TabPFN's transformer into the imaging and tabular encoders, aligning their representations with the inductive biases of TabPFN.

\section{Method}\label{sec:method}

An overview of M$^2$PFN is shown in Fig.~\ref{fig:overview}. We first formulate multimodal in-context prediction (Sec.~\ref{sec:method:formulation}), then describe the alignment-based representation learner (Sec.~\ref{sec:method:repr}), the differentiable TabPFN prediction head (Sec.~\ref{sec:method:tabpfn}), and the gated tabular shortcut (Sec.~\ref{sec:method:fusion}). The complete training objective is given in Sec.~\ref{sec:method:objective}.

\subsection{Problem Formulation}\label{sec:method:formulation}

Subject $i$ is represented by $\mathbf{x}_i=(x_i^{\mathrm{img}},x_i^{\mathrm{tab}})$, where $x_i^{\mathrm{img}}\in\mathbb{R}^{1\times128\times128\times128}$ is a 3D MRI volume and $x_i^{\mathrm{tab}}\in\mathbb{R}^{d}$ contains continuous and ordinal-encoded categorical variables. The task target is denoted by $y_i$: for diagnosis, $y_i\in\{0,1,2\}$ corresponds to $\{\mathrm{CN},\mathrm{MCI},\mathrm{AD}\}$; for cognitive-score regression, $y_i\in\mathbb{R}$ (\S\ref{sec:exp:reg}). We additionally denote the diagnostic category by $c_i\in\{0,1,2\}$ for the prototype-alignment loss below. Thus, $c_i=y_i$ for classification, whereas $c_i$ is an auxiliary diagnosis label for regression. Given a query subject $q$ and a labeled support set $\mathcal{D}_{S}=\{(\mathbf{x}_j,y_j)\}_{j\in\mathcal{S}}$, both tasks are formulated as
\begin{equation}
p_{\Theta_{\mathrm{PFN}}}\!\left(
y_q\,\middle|\,\tilde h_q,
\{(\tilde h_j,y_j)\}_{j\in\mathcal{S}}
\right),
\label{eq:icl-formulation}
\end{equation}
where $\tilde h_i\in\mathbb{R}^{D'}$ is the learned multimodal representation supplied to TabPFN, defined in Sec.~\ref{sec:method:tabpfn}, and $\Theta_{\mathrm{PFN}}$ denotes the pretrained TabPFN parameters. For classification, $\hat y_q=\arg\max_{c}p(y_q=c\mid\cdot)$; for regression, $\hat y_q$ is the posterior mean of TabPFN's discretized target distribution. Because the support set enters only as context, $\mathcal{D}_{S}$ can be replaced at inference without updating the model parameters, enabling the external support-swap evaluation in \S\ref{sec:exp:extval}. Our inference path is differentiable with respect to $\tilde h_q$ and hence the query inputs. The PFN weights $\Theta_{\mathrm{PFN}}$ remain frozen, while gradients optimize the modality encoders, disentangler, projection, and fusion gate.

\subsection{Alignment-Based Multimodal Representation Learning}\label{sec:method:repr}

\textbf{Per-modality encoders.}
The visual encoder $E_v(\cdot;\theta_v)$ maps $x_i^{\mathrm{img}}$ to $u_i^v\in\mathbb{R}^{H}$, and the tabular encoder $E_t(\cdot;\theta_t)$ maps $x_i^{\mathrm{tab}}$ to $u_i^t\in\mathbb{R}^{H}$, with $H=768$ and superscripts $v$ and $t$ denoting visual and tabular modalities. Specifically, $E_v$ is a MONAI~\cite{cardoso2022monai} 3D ResNet-18~\cite{he2016deep} initialized from MedicalNet~\cite{chen20193dresnet}. Global average pooling produces a 512-dimensional descriptor, followed by Linear$+$LayerNorm$+$GELU to obtain $u_i^v$. The FT-Transformer~\cite{gorishniy2021ftt} used for $E_t$ embeds each continuous or categorical variable as a token, prepends a learnable \texttt{[CLS]} token, and applies three width-$H$ transformer blocks with eight attention heads. Its \texttt{[CLS]} output is mapped by Linear$+$RMSNorm to $u_i^t$.

\textbf{Shared/specific disentanglement.}
For each modality $m\in\{v,t\}$, two parallel heads transform $u_i^m$ into a shared representation $s_i^m=S_m(u_i^m)\in\mathbb{R}^{D_s}$ and a modality-specific representation $p_i^m=P_m(u_i^m)\in\mathbb{R}^{D_p}$. Each head comprises Linear$+$LayerNorm$+$GELU, with $D_s=D_p=576$ (Fig.~\ref{fig:overview}(b)); their parameters are collectively denoted by $\phi$. The heads have the same architecture but separate parameters. Their functional roles are induced by the alignment objective: only $s_i^v$ and $s_i^t$ are directly aligned, whereas $p_i^v$ and $p_i^t$ are optimized only through the downstream task loss.

\textbf{Class-prototype alignment loss.}
Instead of aligning individual MRI--tabular pairs with an InfoNCE objective~\cite{radford2021learning,li2021align}, we align class prototypes within the current query minibatch $\mathcal{B}$ (Fig.~\ref{fig:overview}(c)). Let
$\mathcal{B}_c=\{i\in\mathcal{B}:c_i=c\}$ and
$\mathcal{C}_{\mathcal{B}}=\{c:\mathcal{B}_c\neq\varnothing\}$.
The modality-$m$ prototype for category $c$ is
\begin{equation}
\bar s_c^m
=\frac{1}{|\mathcal{B}_c|}
\sum_{i\in\mathcal{B}_c}s_i^m,
\end{equation}
and the alignment loss is
\begin{equation}
\mathcal{L}_{\mathrm{align}}
=\frac{1}{|\mathcal{C}_{\mathcal{B}}|}
\sum_{c\in\mathcal{C}_{\mathcal{B}}}
\left[
1-\cos\!\left(\bar s_c^v,\bar s_c^t\right)
\right],
\label{eq:align}
\end{equation}
where $\cos(a,b)=a^\top b/(\|a\|_2\|b\|_2)$. The number of alignment targets is therefore bounded by the number of diagnostic categories rather than the number of subjects, limiting memorization of individual cross-modal pairs. Minimizing Eq.~\eqref{eq:align} encourages the shared heads to encode diagnosis-consistent cross-modal information, while the specific heads remain free to retain complementary task-relevant information.

\textbf{Concatenated representation.}
Following the ordering in Fig.~\ref{fig:overview}(a), the modality-specific and shared representations are concatenated as
\begin{equation}
h_i^{C}
=
[\,p_i^v;\,s_i^v;\,s_i^t;\,p_i^t\,]
\in\mathbb{R}^{2(D_s+D_p)}
=
\mathbb{R}^{2304},
\label{eq:fused-embedding}
\end{equation}
where the superscript $C$ denotes concatenation.

\subsection{Differentiable ICL Classification and Regression}\label{sec:method:tabpfn}

A learnable Linear$+$LayerNorm projection
\[
\mathrm{Proj}_{\psi}:
\mathbb{R}^{2(D_s+D_p)}
\rightarrow
\mathbb{R}^{D'}
\]
maps the concatenated representation to a compact fused representation:
\begin{equation}
h_i^{\mathrm{fused}}
=
\mathrm{Proj}_{\psi}(h_i^{C}),
\qquad D'=512.
\label{eq:fused-projection}
\end{equation}
Given the fused representation $h_q^{\mathrm{fused}}$ of query subject $q$ and a cached support context, the multimodal ICL branch computes
\begin{equation}
z_{\mathrm{mm},q}
=
\mathrm{TabPFN}\!\left(
h_q^{\mathrm{fused}},
\left\{
\bigl(
\operatorname{sg}(h_j^{\mathrm{fused}}),y_j
\bigr)
\right\}_{j\in\mathcal{S}};
\Theta_{\mathrm{PFN}}
\right),
\label{eq:tabpfn-head}
\end{equation}
where $\operatorname{sg}(\cdot)$ denotes the stop-gradient operation applied to the support representations, as illustrated in Fig.~\ref{fig:overview}(a), and $\Theta_{\mathrm{PFN}}$ denotes the frozen TabPFN parameters. For classification,
$z_{\mathrm{mm},q}\in\Delta^{K-1}$ is a probability vector over
$K=3$ diagnostic classes. For regression, the TabPFN classifier is replaced by a TabPFN regressor, and
$z_{\mathrm{mm},q}\in\mathbb{R}$ denotes the posterior mean of its discretized predictive distribution in standardized target space. The projection is optimized jointly with the modality encoders and disentanglement heads, allowing the task loss to retain the feature directions most useful to the frozen ICL engine.

\textbf{Differentiable inference.}
We bypass TabPFN's non-differentiable feature preprocessing and retain differentiable PyTorch z-normalization before the transformer, as illustrated in Fig.~\ref{fig:overview}(d). The transformer forward pass remains differentiable even though $\Theta_{\mathrm{PFN}}$ is frozen. Consequently, the task gradient follows
$z_{\mathrm{mm},q}
\rightarrow h_q^{\mathrm{fused}}
\rightarrow h_q^{C}
\rightarrow \{u_q^v,u_q^t\},$
and updates the trainable front end. No gradient is propagated through the detached support cache. Freezing $\Theta_{\mathrm{PFN}}$ preserves its meta-learned in-context prior, while adapting its input representation to multimodal AD data, analogous to continuous prompt tuning.

\textbf{Per-epoch support refresh.}
At epoch $e$, the training set is randomly partitioned into a support subset $\mathcal{S}^{(e)}$ and a disjoint query subset $\mathcal{Q}^{(e)}$. Support subjects are encoded once using the current front-end parameters, detached, and reused for all query updates in that epoch. For classification, $1{,}400$ of $1{,}556$ training subjects form the support, leaving $156$ queries processed in ten batches of at most 16. For MMSE regression, $859$ of $875$ subjects form the support, and the remaining 16 queries constitute one ICL episode and one optimizer update. Support caching avoids repeated 3D MRI encoding, although the cached representations become slightly stale as the front end is updated within an epoch.

\subsection{Gated TabPFN Shortcut}\label{sec:method:fusion}

The multimodal branch uses both MRI and tabular representations. As shown in Fig.~\ref{fig:overview}(a), we add a frozen tabular-only TabPFN shortcut and blend its output with $z_{\mathrm{mm}}$ through a learnable scalar gate. The shortcut provides a calibrated tabular reference, while the multimodal branch learns complementary information from MRI and tabular features.

\textbf{TabPFN shortcut.}
Let $z_{\mathrm{tab},i}\in\Delta^{C-1}$ denote the frozen tabular-only TabPFN prediction for classification. For validation and test subjects, its context is a random subset of the training data ($1{,}400$ of $1{,}556$ subjects). For training subjects, $z_{\mathrm{tab},i}$ is precomputed by stratified $K$-fold out-of-fold inference with $K=5$: each held-out fold is predicted using the other $K-1$ folds as support, so the subject being predicted never appears in its own context. The cache
$\{z_{\mathrm{tab},i}\}_{i\in\mathcal{D}_{\mathrm{train}}}$
is generated once per random seed and retrieved by subject index during training. For regression, $z_{\mathrm{tab},i}\in\mathbb{R}$ is the corresponding posterior mean in standardized target space.

\textbf{Learnable gating.}
For each subject, the two branches are combined by
\begin{equation}
z_{\mathrm{final},i}
=\alpha z_{\mathrm{mm},i}
 +(1-\alpha)z_{\mathrm{tab},i},
\qquad
\alpha=\sigma(\rho),
\label{eq:fusion}
\end{equation}
where $\rho\in\mathbb{R}$ is trainable, $\sigma$ is the sigmoid, and $\rho$ is initialized to zero so that $\alpha=0.5$. For classification, the three $z$ variables are probability vectors; for regression, they are standardized scalars.

Because the shortcut prediction is frozen and detached,
\begin{equation}
\frac{\partial z_{\mathrm{final},i}}
     {\partial h_i^{\mathrm{fused}}}
=
\alpha\,
\frac{\partial z_{\mathrm{mm},i}}
     {\partial h_i^{\mathrm{fused}}}.
\end{equation}
Thus, representation gradients pass only through the multimodal branch, whereas $\rho$ is updated by the final task loss and learns the relative contribution of the two predictions.

\subsection{Training Objective}\label{sec:method:objective}

Let
$\Omega=\{\theta_v,\theta_t,\phi,\psi,\rho\}$
collect the trainable parameters of the two encoders, four disentangler heads, projection, and gate, respectively. For a query minibatch $\mathcal{B}$, we optimize
\begin{equation}
\mathcal{L}(\Omega)
=\mathcal{L}_{\mathrm{task}}
+\lambda_{\mathrm{align}}
 \mathcal{L}_{\mathrm{align}},
\label{eq:total-loss}
\end{equation}
where
\begin{equation}
\mathcal{L}_{\mathrm{task}}
=
\begin{cases}
-\dfrac{1}{|\mathcal{B}|}
 \displaystyle\sum_{i\in\mathcal{B}}
 \log [z_{\mathrm{final},i}]_{y_i},
& \text{classification},\\[8pt]
\dfrac{1}{|\mathcal{B}|}
 \displaystyle\sum_{i\in\mathcal{B}}
 \mathrm{SmoothL1}_{\beta}
 \!\left(z_{\mathrm{final},i}-y_i^z\right),
& \text{regression},
\end{cases}
\label{eq:task-loss}
\end{equation}
and
$y_i^z=(y_i-\mu_y)/\sigma_y$
is the regression target standardized using training-set statistics. For residual $e$,
\begin{equation}
\mathrm{SmoothL1}_{\beta}(e)
=
\begin{cases}
\dfrac{e^2}{2\beta},
& |e|<\beta,\\[4pt]
|e|-\dfrac{\beta}{2},
& \text{otherwise}.
\end{cases}
\label{eq:smoothl1}
\end{equation}
All TabPFN parameters $\Theta_{\mathrm{PFN}}$ remain fixed. We use $\lambda_{\mathrm{align}}=0.1$ and $\beta=1$ unless stated otherwise.

\section{Experiments}\label{sec:experiments}

\subsection{Dataset and Preprocessing}\label{sec:exp:dataset}

Experiments are conducted on three public datasets: the Alzheimer's Disease Neuroimaging Initiative (ADNI)~\cite{mueller2005alzheimer,jack2008alzheimer}, the Standardized Centralized Alzheimer's \& Related Dementias Neuroimaging (SCAN) initiative~\cite{scan2020}, and the Open Access Series of Imaging Studies 3 (OASIS-3)~\cite{lamontagne2019oasis}. All T1-weighted MRI volumes are processed with FreeSurfer~\cite{fischl2012freesurfer}, and the resulting skull-stripped brain images are resampled to $128\!\times\!128\!\times\!128$. The accompanying tabular modality comprises 172 features spanning medical history, neurological examination (excluding cognitive tests), demographics, vital signs, APOE genotype, and CSF biomarkers. 

The three cohorts differ in tabular completeness. ADNI provides the full 172-feature set, whereas OASIS-3 and SCAN provide only 7 and 27 features, respectively. The seven features shared by both cohorts are sex, age, education, handedness, ethnicity, and the two APOE alleles; SCAN additionally supplies race and marital status, three vital signs, nine per-system medical-history indicators, three neurological-exam signs, and CSF amyloid and tau values derived from a coarse two-level status. Available features are coded into their corresponding ADNI slots, and how a missing value is filled depends on the feature type rather than on the cohort: continuous features are mean-imputed, and categorical features are assigned a reserved index whose embedding is learned with the rest of the model. No architectural change is needed, so the same frozen network consumes all three cohorts.

A subject enters the classification cohort if a FreeSurfer-reconstructed baseline T1 and a baseline diagnosis that maps onto CN/MCI/AD are both available. In ADNI this retains the $2240$ subjects that have both, out of $2962$ with a baseline diagnosis. In OASIS-3, the label is the clinician diagnosis recorded at the baseline Uniform Data Set visit, which leaves $1163$ of $1195$ subjects; the $32$ dropped either carry no diagnosis or carry a non-AD dementia with no counterpart in a three-way CN/MCI/AD task. In SCAN, the label is the NACC Uniform Data Set diagnosis~\cite{beekly2007nacc}, and the $34$ of $1151$ subjects coded impaired-but-not-MCI are dropped for the same reason, leaving $1117$. The regression cohort is the subset of the classification cohort carrying a valid baseline cognitive test score for ADNI ($1250$ of $2240$) and SCAN ($1097$ of $1117$), whereas for OASIS-3 it comprises all $1195$ subjects with an MMSE score, including the $32$ that the three-way diagnosis excludes. The target is the Mini-Mental State Examination (MMSE) for ADNI and OASIS-3 and the Montreal Cognitive Assessment (MoCA) for SCAN. ADNI is partitioned into subject-disjoint training, validation, and test sets following a $70/15/15$ split ($1556/333/351$ subjects for classification, $875/185/190$ for regression). All models are trained exclusively on ADNI; OASIS-3 and SCAN are held out and used solely as external test sets. Detailed cohort statistics are reported in Table~\ref{tab:dataset_statistics}, where rows describe the classification cohort.

\begin{table}[th!]
\centering
\caption{Characteristics of the three study cohorts, broken down by diagnostic group. Age is mean$_{\pm\text{std}}$; Sex is shown in female/male; APOE $\varepsilon4^{+}$ is carriers over genotyped subjects.}
\label{tab:dataset_statistics}
\footnotesize
\setlength{\tabcolsep}{2pt}
\begin{tabular}{lllll}
\hline
 & $n$ & Age (Year) & Sex (F/M) & APOE $\varepsilon4^{+}$ \\
\hline
\multicolumn{5}{l}{\textbf{ADNI}} \\
\quad CN  & $1000$ & $71.9_{\pm 7.7}$ & $616/384$ & $223/726$ \\
\quad MCI & $770$  & $73.0_{\pm 7.9}$ & $354/416$ & $258/610$ \\
\quad AD  & $470$  & $74.8_{\pm 7.9}$ & $217/253$ & $279/413$ \\
\quad All & $2240$ & $72.9_{\pm 7.9}$ & $1187/1053$ & $760/1749$ \\
\hline
\multicolumn{5}{l}{\textbf{OASIS-3}} \\
\quad CN  & $855$  & $67.0_{\pm 9.2}$ & $497/358$ & $299/850$ \\
\quad MCI & $95$   & $71.8_{\pm 6.2}$ & $49/46$ & $41/94$ \\
\quad AD  & $213$  & $74.0_{\pm 7.8}$ & $93/120$ & $133/211$ \\
\quad All & $1163$ & $68.7_{\pm 9.2}$ & $639/524$ & $473/1155$ \\
\hline
\multicolumn{5}{l}{\textbf{SCAN}} \\
\quad CN  & $756$  & $71.9_{\pm 8.3}$ & $502/254$ & $247/709$ \\
\quad MCI & $180$  & $74.2_{\pm 7.0}$ & $87/93$ & $70/152$ \\
\quad AD  & $181$  & $72.8_{\pm 8.5}$ & $99/82$ & $88/140$ \\
\quad All & $1117$ & $72.5_{\pm 8.2}$ & $688/429$ & $405/1001$ \\
\hline
\end{tabular}

\end{table}

\begin{table*}[t]
\centering
\caption{Performance on the ADNI test set. Img/Tab indicates which modalities the method consumes.}
\label{tab:main}
\renewcommand{\arraystretch}{0.9}
\resizebox{0.8\textwidth}{!}{%
\begin{tabular}{l|cc|ccccc}
\hline
\textbf{Method} & \textbf{Img} & \textbf{Tab} & \textbf{Acc.\ (\%)} & \textbf{F1 (\%)} & \textbf{AUC (\%)} & \textbf{Sens.\ (\%)} & \textbf{Spec.\ (\%)} \\ \hline
\multicolumn{8}{l}{\textit{Unimodal Baselines}}\\\hline
ResNet~\cite{he2016deep} & \checkmark &            & $46.91_{\pm 3.90}$ & $39.27_{\pm 3.17}$ & $69.70_{\pm 2.17}$ & $46.55_{\pm 3.22}$ & $71.96_{\pm 2.37}$ \\
DenseNet~\cite{huang2017densely} & \checkmark &            & $46.06_{\pm 4.89}$ & $39.43_{\pm 4.33}$ & $68.55_{\pm 2.22}$ & $45.39_{\pm 3.85}$ & $72.20_{\pm 3.54}$ \\
SE-ResNet~\cite{hu2018squeeze} & \checkmark &            & $46.91_{\pm 4.78}$ & $43.27_{\pm 3.51}$ & $62.86_{\pm 1.04}$ & $44.34_{\pm 0.50}$ & $71.31_{\pm 0.94}$ \\
SwinUNETR~\cite{he2023swinunetr} & \checkmark &            & $48.53_{\pm 1.65}$ & $42.32_{\pm 1.34}$ & $64.77_{\pm 0.96}$ & $43.29_{\pm 2.85}$ & $71.89_{\pm 0.51}$ \\
XGBoost~\cite{chen2016xgboost} &            & \checkmark & $57.68_{\pm 0.26}$ & $54.51_{\pm 0.39}$ & $74.83_{\pm 0.23}$ & $53.90_{\pm 0.36}$ & $77.40_{\pm 0.09}$ \\
AutoGluon~\cite{erickson2020autogluon} &            & \checkmark & $60.49_{\pm 1.52}$ & $58.01_{\pm 0.70}$ & $77.77_{\pm 0.18}$ & $57.96_{\pm 1.05}$ & $79.01_{\pm 0.62}$ \\
TabPFN-v2.5~\cite{hollmann2025accurate} &            & \checkmark & $61.57_{\pm 0.39}$ & $59.19_{\pm 0.25}$ & $78.84_{\pm 0.04}$ & $58.64_{\pm 0.20}$ & $78.20_{\pm 0.20}$ \\
\hline
\multicolumn{8}{l}{\textit{Multimodal Baselines}}\\\hline
InterpretCC~\cite{swamy2024intrinsic} & \checkmark & \checkmark & $61.88_{\pm 1.77}$ & $61.51_{\pm 2.37}$ & $80.52_{\pm 0.32}$ & $63.65_{\pm 2.67}$ & $80.44_{\pm 1.14}$ \\
MoE++~\cite{jin2024moe++} & \checkmark & \checkmark & $60.23_{\pm 2.10}$ & $60.32_{\pm 1.84}$ & $79.95_{\pm 0.70}$ & $61.76_{\pm 1.29}$ & $79.61_{\pm 0.82}$ \\
SwitchGate~\cite{fedus2022switch} & \checkmark & \checkmark & $60.23_{\pm 2.41}$ & $59.43_{\pm 3.33}$ & $79.71_{\pm 0.74}$ & $62.12_{\pm 1.96}$ & $79.55_{\pm 1.25}$ \\
mmFormer~\cite{zhang2022mmformer} & \checkmark & \checkmark & $51.34_{\pm 3.42}$ & $49.84_{\pm 4.54}$ & $69.68_{\pm 4.02}$ & $50.76_{\pm 4.48}$ & $74.91_{\pm 1.68}$ \\
MulT~\cite{tsai2019multimodal} & \checkmark & \checkmark & $60.63_{\pm 1.18}$ & $60.39_{\pm 0.85}$ & $80.50_{\pm 0.75}$ & $62.54_{\pm 0.95}$ & $79.64_{\pm 0.53}$ \\
Flex-MoE~\cite{yun2024flex} & \checkmark & \checkmark & $62.91_{\pm 1.85}$ & $62.26_{\pm 1.93}$ & $80.13_{\pm 0.47}$ & \underline{$63.94_{\pm 1.59}$} & $80.73_{\pm 1.10}$ \\
I2MoE~\cite{xin2025i2moe} & \checkmark & \checkmark & $61.42_{\pm 5.88}$ & $60.49_{\pm 6.37}$ & $79.23_{\pm 0.97}$ & $63.14_{\pm 3.90}$ & $80.36_{\pm 2.09}$ \\
MMPFN~\cite{kim2026multimodalpfn} & \checkmark & \checkmark & \underline{$64.50_{\pm 0.80}$} & \underline{$62.85_{\pm 0.95}$} & \underline{$81.14_{\pm 0.45}$} & $63.33_{\pm 0.94}$ & \underline{$81.04_{\pm 0.46}$} \\
\hline
\textbf{M$^2$PFN (ours)}                & \checkmark & \checkmark & $\mathbf{67.18_{\pm 1.13}}$ & $\mathbf{65.55_{\pm 1.26}}$ & $\mathbf{82.21_{\pm 0.54}}$ & $\mathbf{65.71_{\pm 1.44}}$ & $\mathbf{82.33_{\pm 0.64}}$ \\
\hline
\end{tabular}%
}
\end{table*}

\subsection{Implementation Details}\label{sec:exp:impl}
Our model uses the TabPFN-v2.5 checkpoint~\cite{hollmann2025accurate} for both the multimodal in-context head and the tabular shortcut in M$^2$PFN. Per-layer activation recomputation is enabled within the TabPFN backbone to improve memory efficiency. For classification, we use a learning rate of $5\!\times\!10^{-5}$, weight decay $10^{-4}$, batch size $16$, $175$ epochs, and a linear warmup ratio of $0.1$. The regression instantiation (\S\ref{sec:exp:reg}) adopts the same hyperparameter configuration, except that it is trained for $1200$ epochs with a batch size of $1$. Both instantiations are trained and evaluated with $5$ random seeds, and we report the mean $\pm$ standard deviation across seeds; a single seed requires approximately $5.5$ hours for classification and $23$ hours for regression. Statistical comparisons are performed with a paired one-sided Student's $t$-test on the per-seed metrics. We report Accuracy, macro F1, macro AUC (one-vs-rest), Sensitivity, and Specificity for classification, and MAE, Pearson $r$, and $R^2$ for regression, all computed on the held-out test set.

\subsection{Baselines}\label{sec:exp:baselines}
For baselines, we consider three families of methods: unimodal models operating on the MRI volume, unimodal models operating on the tabular data, and multimodal fusion models. All baselines use the standard public implementation of their respective architectures and are trained and evaluated on the same splits. To ensure a fair comparison on the imaging side, any multimodal baseline whose released code does not provide a $3$D image encoder is equipped with the same MedicalNet-pretrained $3$D ResNet-$18$ vision encoder used by M$^2$PFN, fine-tuned end-to-end, while retaining its own tabular encoder. For every baseline, we adopt the default hyperparameters specified in the published code. The full list of baselines is as follows:
\begin{itemize}
\item \emph{Unimodal MRI:} ResNet-3D~\cite{he2016deep}, DenseNet-3D~\cite{huang2017densely}, SE-ResNet-3D~\cite{hu2018squeeze}, SwinUNETR~\cite{he2023swinunetr}.
\item \emph{Unimodal tabular:} XGBoost~\cite{chen2016xgboost}, AutoGluon~\cite{erickson2020autogluon}, TabPFN-v2.5~\cite{hollmann2025accurate}.
\item \emph{Multimodal fusion:} InterpretCC~\cite{swamy2024intrinsic}, MoE++~\cite{jin2024moe++}, SwitchGate~\cite{fedus2022switch}, mmFormer~\cite{zhang2022mmformer}, MulT~\cite{tsai2019multimodal}, Flex-MoE~\cite{yun2024flex}, I2MoE~\cite{xin2025i2moe}, MMPFN~\cite{kim2026multimodalpfn}.
\end{itemize}

\subsection{Classification on ADNI}\label{sec:exp:main}
We perform a three-way diagnostic classification experiment on the ADNI dataset. Table~\ref{tab:main} summarizes the ADNI test-set results. M$^2$PFN achieves the best performance across all evaluation metrics, including accuracy ($67.18\%$), macro-F1 ($65.55\%$), macro-AUC ($82.21\%$), sensitivity ($65.71\%$), and specificity ($82.33\%$). It consistently outperforms both unimodal and multimodal baselines, demonstrating the benefit of integrating MRI with tabular biomarkers. Among the multimodal methods, MMPFN is the strongest competitor, while Flex-MoE and InterpretCC also improve upon tabular-only models but remain behind M$^2$PFN. Notably, both MMPFN and M$^2$PFN use the TabPFN ICL engine; the key difference is that MMPFN relies on frozen image embeddings, whereas M$^2$PFN optimizes the imaging encoder end-to-end through the ICL transformer. The statistically significant improvement over MMPFN therefore highlights the effectiveness of end-to-end prompt tuning rather than in-context learning alone.

\begin{table}[t]
\centering
\caption{Component ablation on the ADNI test set. Each row removes one mechanism from the full recipe.}
\label{tab:ablation}
\resizebox{0.6\textwidth}{!}{%
\begin{tabular}{l|ccc}
\hline
\textbf{Variant} & \textbf{Acc.\ (\%)} & \textbf{F1 (\%)} & \textbf{AUC (\%)} \\ \hline
Full M$^2$PFN & $\mathbf{67.18_{\pm 1.13}}$ & $\mathbf{65.55_{\pm 1.26}}$ & $\mathbf{82.21_{\pm 0.54}}$ \\ \hline
w/o Disentangled Alignment (DA) & $65.70_{\pm 1.18}$ & $64.36_{\pm 1.72}$ & $81.41_{\pm 0.70}$ \\
w/o End-to-end Training (E2E) & $65.41_{\pm 1.47}$ & $64.29_{\pm 1.37}$ & $81.72_{\pm 1.12}$ \\
w/o DA \& E2E & $63.93_{\pm 1.52}$ & $62.33_{\pm 1.66}$ & $80.77_{\pm 1.00}$ \\
w/o Gated TabPFN Shortcut & $62.96_{\pm 1.27}$ & $61.00_{\pm 1.00}$ & $79.30_{\pm 0.93}$ \\
w/o ICL (TabPFN $\to$ MLP) & $60.06_{\pm 3.42}$ & $59.32_{\pm 3.74}$ & $76.18_{\pm 2.13}$ \\
\hline
\end{tabular}%
}
\end{table}

\subsection{Ablation Study}\label{sec:exp:ablation}

\begin{figure*}[t!]
\centering
\includegraphics[width=\textwidth]{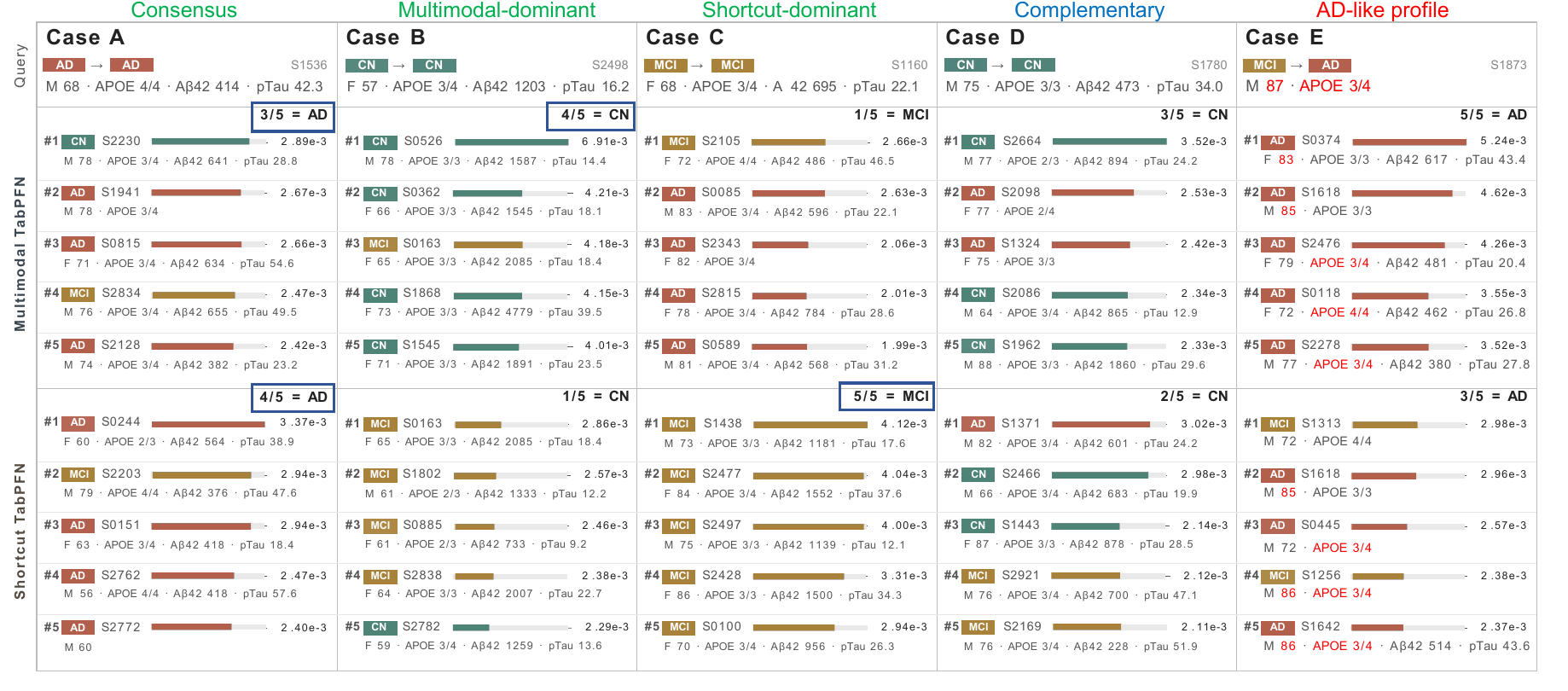}
\caption{Exemplar reference attribution for five representative ADNI test subjects. Each column corresponds to one query subject. The first row summarizes the ground-truth and predicted diagnosis together with selected clinical variables (sex, age, APOE genotype, A$\beta_{42}$, and pTau; missing values indicate unavailable biomarkers). The two panels show the top-5 support subjects retrieved by the multimodal TabPFN and the shortcut TabPFN, respectively. Bar length represents the normalized attention weight within each query, and colors denote diagnostic class (CN, MCI, and AD). Subject identifiers are anonymized (S$n$), and ages above 90 years are reported as $90+$ in accordance with the ADNI Data Use Agreement.}
\label{fig:case-cards}
\end{figure*}

To assess the contribution of each component of M$^2$PFN to the final performance, we conduct an ablation study, reported in Table~\ref{tab:ablation}. The disentangled alignment module consistently improves both predictive performance and training stability, with its removal causing a modest but significant reduction in AUC ($p<0.05$). In contrast, the architectural designs contribute more substantially. End-to-end optimization enables gradients from the frozen ICL engine to guide multimodal representation learning, and jointly removing end-to-end optimization and representation alignment leads to a much larger performance degradation than removing either component alone, indicating that the two are complementary. The frozen TabPFN shortcut further encourages the multimodal branch to learn complementary information beyond what is already captured by the tabular features, leading to richer multimodal representations and better performance. Finally, replacing the frozen TabPFN heads with trainable MLP classifiers produces the largest performance drop, demonstrating that preserving TabPFN's pretrained in-context learning capabilities, rather than a conventional parametric classifier, is the largest source of M$^2$PFN's performance gain.

\subsection{Subject-level Interpretability}
\label{sec:exp:interp}

Within our framework, each query is classified by attending to a labeled
support set. The attention assigned to each support subject therefore
provides an example-level measure of its contribution to the query
prediction. For query $q$, we obtain the multimodal attention distribution
$a^{\mathrm{mm}}_{q,\cdot}\in\Delta^{|\mathcal{S}|-1}$ by extracting the
softmax attention weights from each TabPFN layer and head, averaging them
over layers and heads, and renormalizing the resulting vector over the
support set $\mathcal{S}$. We similarly obtain
$a^{\mathrm{tab}}_{q,\cdot}\in\Delta^{|\mathcal{S}|-1}$ from the
tabular-only TabPFN shortcut. Both branches use the same indexed support
set $\mathcal{S}$.

Because the final prediction is
\begin{equation}
z_{\mathrm{final},q}
=
\alpha z_{\mathrm{mm},q}
+
(1-\alpha)z_{\mathrm{tab},q},
\end{equation}
we define a branch-weighted unified attribution by
\begin{equation}
a^{\mathrm{uni}}_{q,j}
=
\alpha a^{\mathrm{mm}}_{q,j}
+
(1-\alpha)a^{\mathrm{tab}}_{q,j},
\qquad j\in\mathcal{S}.
\label{eq:unified-attribution}
\end{equation}
Since both branch-specific attention vectors lie in the probability
simplex and $\alpha\in[0,1]$, their convex combination satisfies
$a^{\mathrm{uni}}_{q,\cdot}\in\Delta^{|\mathcal{S}|-1}$. Consequently,
entropy, top-$k$, and effective-support-size analyses remain applicable.
The value $a^{\mathrm{uni}}_{q,j}$ represents the relative
attention-based attribution of support subject $j$ to query $q$.

Figure~\ref{fig:case-cards} visualizes the attribution for five
representative test subjects by showing the five most highly attended
support subjects from the multimodal and shortcut TabPFN branches. Several
representative retrieval patterns emerge. First, Case~A illustrates branch
consensus: both branches predominantly retrieve AD support subjects and
predict AD. Second, Cases~B and C show that the two branches provide
different evidence across query subjects. In Case~B, the multimodal branch
predominantly retrieves CN support subjects, whereas the shortcut branch
favors MCI subjects; their gated combination correctly predicts CN. In
Case~C, the shortcut branch predominantly retrieves MCI support subjects
and offsets the multimodal branch's tendency toward AD. Third, Case~D
illustrates complementary evidence: the multimodal branch retrieves mostly
CN support subjects, whereas the shortcut branch retrieves more MCI and AD
subjects, and their fusion correctly predicts CN. Finally, Case~E presents
a representative failure case in which both branches predominantly retrieve
AD support subjects for an MCI query, resulting in an AD prediction.

These case studies provide insight into the in-context reasoning of
M$^2$PFN. The branch-consensus case suggests that the model predicts by
referencing clinically similar support subjects rather than attending
uniformly over the context. Cases~B--D further indicate that the multimodal
and shortcut branches capture complementary information. Although the
learned gate $\alpha$ is shared across subjects, the branch predictions and
their attended support neighborhoods vary with each query, allowing their
fixed weighted combination to integrate different evidence across subjects.
The misclassified Case~E remains clinically plausible: the query and most
of its highly attended support subjects exhibit AD-like clinical profiles,
including advanced age and APOE-$\varepsilon4$ positivity. This suggests
that the prediction follows clinically meaningful evidence in the support
set rather than arbitrary model behavior. Together, these examples show
that M$^2$PFN provides subject-level interpretability by exposing the
support subjects most strongly associated with each prediction.

\begin{figure}[t]
\centering
\includegraphics[width=0.6\linewidth]{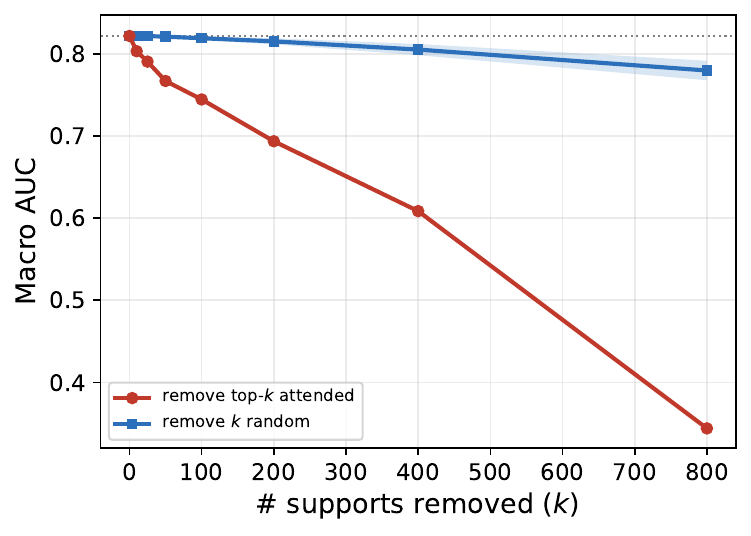}
\caption{Deletion-curve faithfulness test on all 351 ADNI test queries. Removing the top-k most-attended unified supports (red) collapses macro-AUC far faster than removing k random supports (blue; mean$\pm$std over 5 draws). Dotted line: the unablated baseline.}
\label{fig:deletion}
\end{figure}

In addition to the interpretability experiment, we also validate that this attention reflects causal influence with a leave-one-support-out (LOO) probe. For each query, we drop the top-$50$ unified-attention support subjects one at a time (plus $50$ random controls) and measure the induced change $|\Delta z_{\mathrm{final}}|$ in the predicted-class posterior, denoted as LOO influence. On a class-stratified $30$-query subset, the ratio of mean top-$50$ influence to mean control is $4.52\!\pm\!1.29$ (median $4.38$), with $100\%$ of queries above unity, and the per-query Spearman correlation between attention and LOO influence is $0.639$ (median $0.647$), indicating that highly attended support subjects are also those whose removal most strongly perturbs the prediction. We further validate the attention ranking with a standard deletion-curve faithfulness test on all $351$ queries (Fig.~\ref{fig:deletion}), showing that ablating the top-$k$ most-attended supports degrades task performance far faster than ablating $k$ random supports (mean over 5 draws). Removing the top-800 collapses macro-AUC from 0.822 to 0.344, whereas the same number of supports removed at random leaves macro-AUC slightly degraded (0.780), with the gap widening monotonically in $k$. Attention therefore focuses each prediction on a small subset of support examples that carries the most informative causal evidence.

\begin{table}[t]
\centering
\caption{Test-set regression of baseline MMSE on the $1{,}250$-subject MMSE sub-cohort of ADNI.}
\label{tab:reg-main}
\resizebox{0.6\textwidth}{!}{%
\begin{tabular}{l|ccc}
\hline
\textbf{Method} & \textbf{MAE} & \textbf{Pearson $r$} & \textbf{$R^{2}$} \\ \hline
InterpretCC~\cite{swamy2024intrinsic} & $1.925_{\pm 0.039}$ & $0.492_{\pm 0.018}$ & $0.225_{\pm 0.021}$ \\
MoE++~\cite{jin2024moe++} & $1.873_{\pm 0.024}$ & $0.540_{\pm 0.017}$ & $0.277_{\pm 0.022}$ \\
SwitchGate~\cite{fedus2022switch} & $1.980_{\pm 0.095}$ & $0.514_{\pm 0.037}$ & $0.117_{\pm 0.101}$ \\
mmFormer~\cite{zhang2022mmformer} & $2.159_{\pm 0.057}$ & $0.275_{\pm 0.059}$ & $0.021_{\pm 0.061}$ \\
MulT~\cite{tsai2019multimodal} & $1.868_{\pm 0.042}$ & $0.556_{\pm 0.015}$ & $0.254_{\pm 0.033}$ \\
Flex-MoE~\cite{yun2024flex} & $1.899_{\pm 0.053}$ & $0.548_{\pm 0.031}$ & $0.256_{\pm 0.049}$ \\
I2MoE~\cite{xin2025i2moe} & $1.949_{\pm 0.025}$ & $0.526_{\pm 0.006}$ & $0.194_{\pm 0.014}$ \\
MMPFN~\cite{kim2026multimodalpfn} & \underline{$1.784_{\pm 0.041}$} & \underline{$0.601_{\pm 0.014}$} & \underline{$0.287_{\pm 0.057}$} \\ \hline
\textbf{M$^2$PFN (ours)} & $\mathbf{1.743_{\pm 0.014}}$ & $\mathbf{0.630_{\pm 0.007}}$ & $\mathbf{0.343_{\pm 0.006}}$ \\
\hline
\end{tabular}%
}
\end{table}

\subsection{Generalization to Cognitive Score Regression}
\label{sec:exp:reg}
To evaluate whether the proposed framework generalizes beyond categorical diagnosis, we instantiate the same architecture for continuous MMSE regression by replacing (i) the TabPFN classifier head with a TabPFN regressor head that emits a scalar mean over the bin-distributional output of TabPFN, and (ii) the cross-entropy task loss with a smooth-$L_1$ (Huber) loss on standardized targets. The disentangler, alignment loss, shortcut branch, and end-to-end training are identical to the classification setup. The target is the baseline Mini-Mental State Examination total score (MMSE; integer scale $0$--$30$, with lower scores indicating greater cognitive impairment), the most widely used cognitive screen in AD research and a standard endpoint for monitoring disease progression. All metrics are reported on the original raw clinical scale (MMSE scores).

Table~\ref{tab:reg-main} summarizes the MMSE regression results on ADNI. M$^2$PFN achieves the best performance across all evaluation metrics, attaining the lowest MAE together with the highest Pearson correlation and coefficient of determination. It consistently outperforms both conventional multimodal fusion methods and the recent ICL-based MMPFN, while exhibiting lower variance across random seeds. Since both MMPFN and M$^2$PFN are built upon the same TabPFN-based ICL paradigm, the improvement can be attributed to our end-to-end optimization through the frozen ICL transformer and the proposed disentangled alignment-based multimodal representation learning, rather than to in-context learning alone. These results further demonstrate that the proposed framework generalizes beyond categorical diagnosis and can effectively support continuous cognitive score prediction with only a task-specific prediction head.

\begin{table*}[t]
\centering
\caption{External CN/MCI/AD diagnosis on OASIS-3 and SCAN, as one-vs-rest AUC per class and macro-averaged.}
\label{tab:ext-cls-disc}
\resizebox{\textwidth}{!}{%
\begin{tabular}{l|cccc|cccc}
\hline
 & \multicolumn{4}{c|}{OASIS-3 (CN/MCI/AD $=855/95/213$)} & \multicolumn{4}{c}{SCAN ($756/180/181$)} \\
 & \multicolumn{4}{c|}{one-vs-rest AUC} & \multicolumn{4}{c}{one-vs-rest AUC} \\
Method & CN & MCI & AD & macro & CN & MCI & AD & macro \\ \hline
InterpretCC~\cite{swamy2024intrinsic} & $0.780_{\pm.015}$ & $0.463_{\pm.024}$ & $0.807_{\pm.015}$ & $0.684_{\pm.012}$ & $0.772_{\pm.010}$ & $0.450_{\pm.009}$ & $0.851_{\pm.009}$ & $0.691_{\pm.008}$ \\
MoE++~\cite{jin2024moe++} & $0.758_{\pm.021}$ & $0.450_{\pm.021}$ & $0.794_{\pm.020}$ & $0.668_{\pm.020}$ & $0.783_{\pm.018}$ & $0.451_{\pm.062}$ & $0.851_{\pm.021}$ & $0.695_{\pm.019}$ \\
SwitchGate~\cite{fedus2022switch} & $0.776_{\pm.030}$ & $0.432_{\pm.043}$ & $0.801_{\pm.018}$ & $0.670_{\pm.019}$ & $0.797_{\pm.015}$ & $0.394_{\pm.019}$ & $\mathbf{0.856_{\pm.015}}$ & $0.683_{\pm.012}$ \\
mmFormer~\cite{zhang2022mmformer} & $0.735_{\pm.024}$ & $0.554_{\pm.033}$ & $0.758_{\pm.012}$ & $0.682_{\pm.015}$ & $0.697_{\pm.029}$ & $0.515_{\pm.026}$ & $0.747_{\pm.047}$ & $0.653_{\pm.020}$ \\
MulT~\cite{tsai2019multimodal} & $0.776_{\pm.013}$ & $0.427_{\pm.041}$ & \underline{$0.809_{\pm.012}$} & $0.671_{\pm.014}$ & $0.765_{\pm.023}$ & $0.457_{\pm.046}$ & $0.850_{\pm.019}$ & $0.691_{\pm.025}$ \\
Flex-MoE~\cite{yun2024flex} & $0.778_{\pm.011}$ & $0.437_{\pm.048}$ & $0.806_{\pm.014}$ & $0.674_{\pm.013}$ & $0.779_{\pm.025}$ & $0.429_{\pm.048}$ & $0.852_{\pm.012}$ & $0.687_{\pm.018}$ \\
I2MoE~\cite{xin2025i2moe} & \underline{$0.787_{\pm.007}$} & $0.466_{\pm.107}$ & $0.774_{\pm.042}$ & $0.676_{\pm.044}$ & $0.780_{\pm.026}$ & $0.404_{\pm.018}$ & $0.841_{\pm.014}$ & $0.675_{\pm.014}$ \\
MMPFN~\cite{kim2026multimodalpfn} (zero-shot) & $0.569_{\pm.048}$ & $0.529_{\pm.021}$ & $0.565_{\pm.062}$ & $0.554_{\pm.042}$ & $0.478_{\pm.029}$ & $0.441_{\pm.023}$ & $0.495_{\pm.031}$ & $0.471_{\pm.024}$ \\
MMPFN~\cite{kim2026multimodalpfn} (support-swap) & $0.768_{\pm.005}$ & \underline{$0.605_{\pm.016}$} & $0.791_{\pm.007}$ & \underline{$0.721_{\pm.006}$} & \underline{$0.819_{\pm.004}$} & $\mathbf{0.735_{\pm.008}}$ & $0.817_{\pm.009}$ & \underline{$0.790_{\pm.004}$} \\ \hline
M$^2$PFN (zero-shot) & $0.783_{\pm.010}$ & $0.520_{\pm.078}$ & $0.807_{\pm.011}$ & $0.703_{\pm.022}$ & $0.762_{\pm.026}$ & $0.527_{\pm.074}$ & $0.810_{\pm.018}$ & $0.700_{\pm.034}$ \\
\textbf{M$^2$PFN (support-swap)} & $\mathbf{0.797_{\pm.007}}$ & $\mathbf{0.607_{\pm.019}}$ & $\mathbf{0.822_{\pm.007}}$ & $\mathbf{0.742_{\pm.010}}$ & $\mathbf{0.837_{\pm.006}}$ & \underline{$0.734_{\pm.009}$} & \underline{$0.853_{\pm.008}$} & $\mathbf{0.808_{\pm.003}}$ \\ \hline
\end{tabular}%
}
\end{table*}

\subsection{External Validation on OASIS-3 and SCAN}
\label{sec:exp:extval}

To test generalization beyond ADNI, we evaluate the \emph{frozen} ADNI checkpoints on \textbf{OASIS-3}~\cite{lamontagne2019oasis} and \textbf{SCAN} with \emph{no gradient updates}, in two modes: \emph{zero-shot}, where external subjects are predicted from the ADNI support set, and \emph{support-swap}, where that support set is replaced by available labeled samples of the external cohort. All baselines and our model receive the same labeled sample as support in each setting and is scored on the same held-out test subjects. A parametric baseline cannot accept labels as input, so we use them for the strongest correction available to its output: a class-prior correction for diagnosis \cite{saerens2002adjusting}, and an affine rescaling for regression. The two ICL-based methods can use them as the support set.

\begin{table*}[t]
\centering
\caption{External regression on OASIS-3 (MMSE) and SCAN (MoCA).}
\label{tab:ext-reg-baselines}
\resizebox{\textwidth}{!}{%
\begin{tabular}{l|ccc|ccc}
\hline
 & \multicolumn{3}{c|}{OASIS-3 (MMSE, same instrument as ADNI)} & \multicolumn{3}{c}{SCAN (MoCA, cross-instrument)} \\
Method & MAE\,$\downarrow$ & $r$\,$\uparrow$ & $R^2$\,$\uparrow$ & MAE\,$\downarrow$ & $r$\,$\uparrow$ & $R^2$\,$\uparrow$ \\ \hline
InterpretCC~\cite{swamy2024intrinsic} & $1.60_{\pm.03}$ & $0.480_{\pm.022}$ & $0.225_{\pm.016}$ & $3.54_{\pm.10}$ & $0.514_{\pm.037}$ & $0.252_{\pm.035}$ \\
MoE++~\cite{jin2024moe++} & $1.60_{\pm.05}$ & $0.480_{\pm.027}$ & \underline{$0.227_{\pm.022}$} & $3.57_{\pm.12}$ & $0.503_{\pm.050}$ & $0.242_{\pm.047}$ \\
SwitchGate~\cite{fedus2022switch} & $1.65_{\pm.06}$ & $0.434_{\pm.076}$ & $0.186_{\pm.056}$ & $3.98_{\pm.08}$ & $0.230_{\pm.127}$ & $0.061_{\pm.039}$ \\
mmFormer~\cite{zhang2022mmformer} & $1.74_{\pm.07}$ & $0.310_{\pm.098}$ & $0.097_{\pm.053}$ & $3.95_{\pm.18}$ & $0.216_{\pm.098}$ & $0.049_{\pm.050}$ \\
MulT~\cite{tsai2019multimodal} & $1.59_{\pm.04}$ & \underline{$0.489_{\pm.033}$} & $\mathbf{0.234_{\pm.027}}$ & $3.63_{\pm.23}$ & $0.475_{\pm.072}$ & $0.218_{\pm.075}$ \\
Flex-MoE~\cite{yun2024flex} & $1.64_{\pm.07}$ & $0.456_{\pm.061}$ & $0.205_{\pm.050}$ & $3.80_{\pm.22}$ & $0.363_{\pm.102}$ & $0.135_{\pm.072}$ \\
I2MoE~\cite{xin2025i2moe} & $1.83_{\pm.03}$ & $0.107_{\pm.052}$ & $0.007_{\pm.017}$ & $4.07_{\pm.05}$ & $0.060_{\pm.092}$ & $0.006_{\pm.015}$ \\
MMPFN~\cite{kim2026multimodalpfn} (zero-shot) & $1.79_{\pm.06}$ & $0.151_{\pm.136}$ & $0.026_{\pm.040}$ & $4.05_{\pm.10}$ & $0.152_{\pm.059}$ & $0.021_{\pm.019}$ \\
MMPFN~\cite{kim2026multimodalpfn} (support-swap) & \underline{$1.50_{\pm.02}$} & $0.437_{\pm.039}$ & $0.156_{\pm.024}$ & \underline{$3.48_{\pm.15}$} & \underline{$0.526_{\pm.020}$} & \underline{$0.272_{\pm.021}$} \\ \hline
M$^2$PFN (zero-shot) & $1.61_{\pm.01}$ & $0.475_{\pm.010}$ & $0.224_{\pm.009}$ & $3.75_{\pm.11}$ & $0.371_{\pm.038}$ & $0.134_{\pm.032}$ \\
\textbf{M$^2$PFN (support-swap)} & $\mathbf{1.47_{\pm.03}}$ & $\mathbf{0.500_{\pm.038}}$ & $0.212_{\pm.031}$ & $\mathbf{3.16_{\pm.12}}$ & $\mathbf{0.643_{\pm.027}}$ & $\mathbf{0.386_{\pm.032}}$ \\ \hline
\end{tabular}%
}
\end{table*}

\textbf{Diagnostic transfer.} As shown in Table~\ref{tab:ext-cls-disc}, all methods maintain reasonable discrimination between CN and AD, whereas distinguishing the minority MCI class remains the primary challenge. With support-set replacement, M$^2$PFN achieves the highest MCI AUC and the best macro-AUC among parametric learning-based methods on both external cohorts, demonstrating the effectiveness of in-context adaptation for cross-cohort diagnosis. Compared with MMPFN, which is also built on TabPFN, M$^2$PFN maintains substantially stronger zero-shot performance before support replacement and benefits further from support-set adaptation, resulting in significantly higher macro-AUC; this indicates that end-to-end multimodal representation learning and in-context learning are complementary for robust cross-cohort diagnosis.

\textbf{Cognitive-score transfer.} As shown in Table~\ref{tab:ext-reg-baselines}, M$^2$PFN achieves the lowest MAE on both OASIS-3 and SCAN, significantly outperforming all baselines. The two cohorts present different transfer scenarios: OASIS-3 shares the same cognitive assessment (MMSE) as ADNI, whereas SCAN requires cross-instrument transfer to MoCA. Consequently, performance differences on OASIS-3 are relatively modest, with several strong baselines achieving comparable correlations. In contrast, SCAN highlights the advantage of in-context adaptation: while rescaling can reduce prediction error, it cannot improve Pearson correlation because the ranking of predictions remains unchanged. By replacing the support set with MoCA-labeled examples, M$^2$PFN substantially improves the correlation and clearly outperforms all competing methods, including the ICL-based MMPFN. These results indicate that effective cross-instrument transfer relies not only on support-set adaptation, but also on learning transferable multimodal representations that remain compatible with the ICL engine.

\section{Conclusion}\label{sec:conclusion}
We presented M$^2$PFN, an end-to-end multimodal framework that repurposes a frozen TabPFN as the prediction engine for Alzheimer's disease analysis by enabling end-to-end optimization through its in-context learning (ICL) transformer. Rather than fine-tuning the foundation model, M$^2$PFN keeps the ICL engine frozen and instead learns multimodal representations that are directly optimized for in-context reasoning through differentiable prompt tuning. A disentangled alignment-based multimodal representation learner and a gated tabular shortcut further bridge heterogeneous MRI and tabular biomarkers while encouraging complementary information learning. Extensive experiments demonstrate that M$^2$PFN achieves state-of-the-art performance on both three-class AD diagnosis and MMSE regression on ADNI, consistently outperforming strong multimodal baselines. More importantly, preserving the frozen ICL engine enables gradient-free support-set replacement at inference, allowing the same model to transfer effectively to independent clinical cohorts without retraining while maintaining superior performance on both classification and cognitive-score prediction tasks. These results demonstrate that combining multimodal representation learning with in-context learning provides an effective paradigm for robust cross-cohort clinical prediction, and highlight the potential of foundation-model-based in-context learning as an alternative to task-specific finetuning for multimodal medical AI.

\section*{Acknowledgments}
This work was supported by the National Institutes of Health (NIH) under grants R01EB022744, R01AG077578, R01AG064584, S10OD032285, U19AG078109, and P30AG066530. This work used data from the Alzheimer's Disease Neuroimaging Initiative (ADNI), OASIS-3, and the Standardized Centralized Alzheimer's and Related Dementias Neuroimaging (SCAN) initiative linked to the National Alzheimer's Coordinating Center (NACC). The NACC database is funded by NIA/NIH Grant U24 AG072122, and the SCAN initiative by NIA/NIH Grant U24 AG067418; NACC and SCAN data are contributed by the NIA-funded Alzheimer's Disease Research Centers (ADRCs).

\bibliographystyle{unsrt}
\bibliography{reference}

\end{document}